\documentclass[12pt]{article}

\usepackage[margin=1in]{geometry}
\usepackage{amsmath,amssymb,amsthm}
\usepackage{graphicx}
\usepackage{booktabs}
\usepackage{multirow}
\usepackage{array}
\usepackage{xcolor}
\usepackage{hyperref}
\usepackage{natbib}
\usepackage{microtype}
\usepackage{caption}
\usepackage{subcaption}
\usepackage{algorithm}
\usepackage{algpseudocode}
\usepackage{enumitem}
\usepackage{float}
\usepackage{url}
\usepackage{fancyvrb}   

\hypersetup{
  colorlinks=true,
  linkcolor=blue!60!black,
  citecolor=blue!60!black,
  urlcolor=blue!60!black
}

\newcommand{\state}{s}
\newcommand{\action}{a}
\newcommand{\reward}{r}
\newcommand{\policy}{\pi}
\newcommand{\discount}{\gamma}

\newcommand{\MAPPO}{\textsc{mappo}}
\newcommand{\QMIX}{\textsc{qmix}}
\newcommand{\IDQN}{\textsc{idqn}}
\newcommand{\DQN}{\textsc{dqn}}

\title{%
  \textbf{Quantum Frog: Emergent Cooperation and Difficulty Scaling}\\
  \textbf{in a Quantized-Time Cooperative Game}\\[6pt]
}

\author{%
  Saad Mankarious\\
  \\
  \texttt{gosaadmakhal@gmail.com}
}

\date{\today}

\begin{document}
\maketitle

\begin{abstract}
We introduce \emph{Quantum Frog}, a two-player cooperative game built on a
novel \emph{quantized-time} mechanic in which the environment advances only
when a player acts.  Inspired by the classic arcade game Frogger, Quantum Frog
requires two frogs to cross an 8$\times$8 grid of traffic and reach the far
side together.  We use reinforcement learning (RL) as an analytical lens to
answer four design questions: (1) how does game difficulty scale with traffic
density, (2) what is the optimal single-agent policy and why, (3) how large is
the cooperation gap between independent and cooperative two-agent play, and
(4) what joint strategy emerges when agents are incentivised to cooperate?
We train agents through five escalating stages, Tabular Q-Learning, Deep
Q-Network (\DQN), Independent \DQN~(\IDQN), and Multi-Agent
Proximal Policy Optimisation (\MAPPO\ with a centralised critic), evaluating
each against traffic densities of one to six cars.  Our key findings are:
(i) the quantized-time mechanic makes a \emph{rush strategy} (moving directly
upward at every step) universally optimal, as time exposure to traffic is
minimised; (ii) adding an uncoordinated second player is harder than sextupling
the traffic for a single expert player; (iii) cooperative training recovers
+32--34 percentage points of joint success rate relative to independent agents
and reduces episode length from $\sim$90 to $\sim$6 steps; and (iv) the
emergent cooperative strategy is synchronised rushing, not complex positional
coordination, illustrating that shared incentives alone suffice to align agents
in time-critical cooperative tasks.  These findings provide concrete,
empirically grounded guidance for the commercial design of Quantum Frog and
offer broader insights into the role of environment mechanics in shaping
multi-agent learning dynamics.
\end{abstract}

\tableofcontents
\newpage

\section{Introduction}
\label{sec:intro}

Designing a multiplayer game is fundamentally a question about incentive
structures: what mechanics encourage players to communicate, to coordinate,
and to find elegant solutions to difficult problems?  Answering such questions
analytically is notoriously hard, human playtesters provide subjective and
noisy signal, and exhaustive human trials are expensive.  Reinforcement
learning agents offer an alternative: a policy trained to optimise a reward
function will expose the true incentive landscape of a game far more
systematically than human play, revealing what behaviours the game
actually rewards.

\begin{figure}
    \centering
    \includegraphics[width=.6\linewidth]{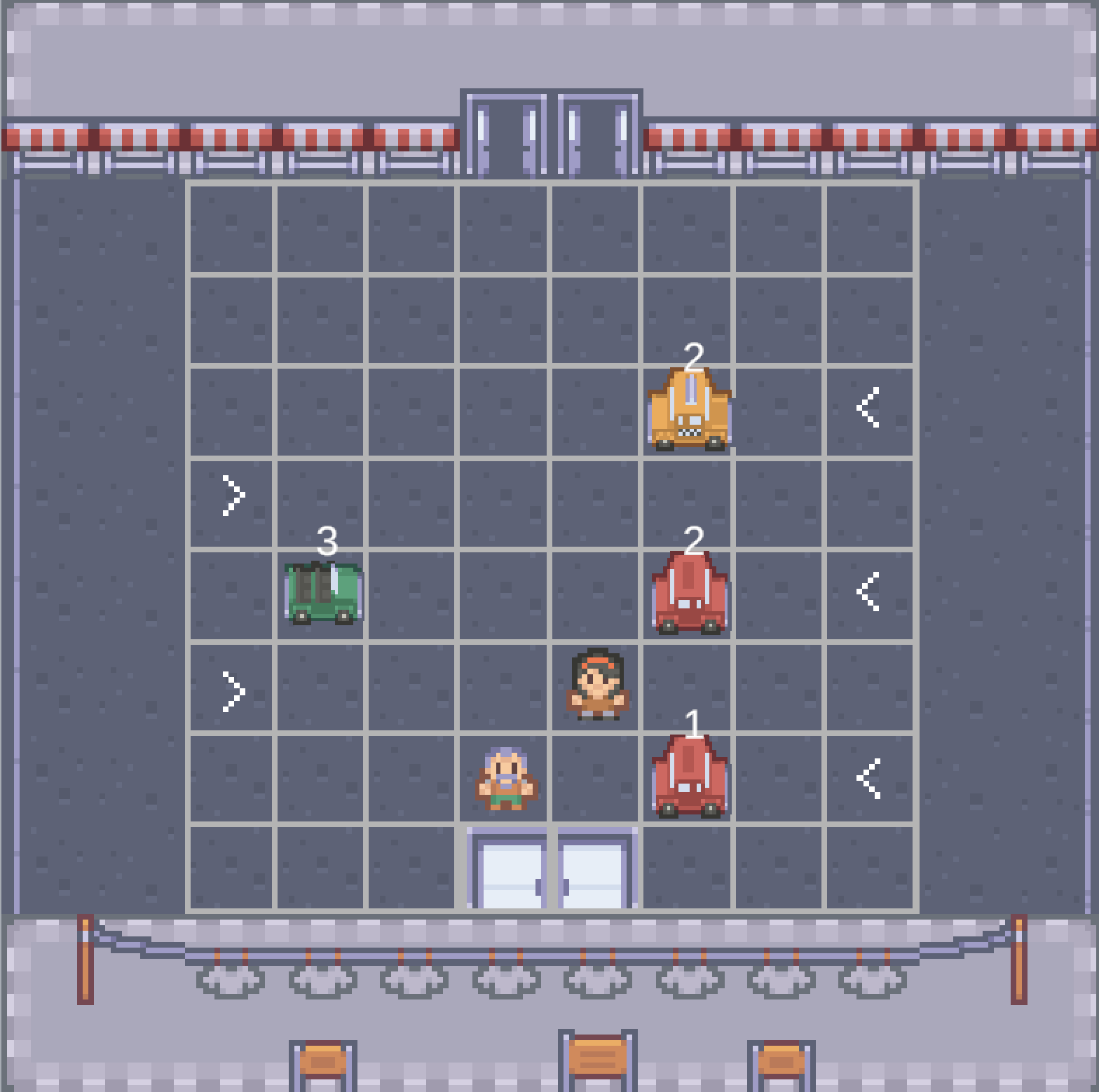}
    \caption{Quantum Frogs Game Representation. Both frogs (represented as boy and girl charachters) begin at the bottom row (row~7);
  four cars move horizontally at speed different speeds and directions.
  The entire board is frozen while the players choose their actions.}
    \label{fig:qf}
\end{figure}

\paragraph{The game.}
Quantum Frog is a two-dimensional, turn-based game played on an 8$\times$8
grid.  Two frogs start at the bottom row and must cross to the top while
avoiding a stream of horizontally moving cars.  The game's defining mechanic
is \emph{quantized time}: the environment is frozen between player decisions,
advancing by exactly one tick whenever a frog acts.  This departs from
continuous-time variants (such as the original Frogger~\citep{konami1981frogger})
in a fundamental way, players can deliberate without penalty and car positions
are perfectly predictable.  The game is cooperative: both frogs must survive
and reach the far side, requiring players to communicate and act jointly.
Figure~\ref{fig:qf} illustrates the board layout and one complete time step.

\paragraph{Motivation.}
The quantized-time mechanic is understudied in game design research.  It
transforms an arcade reflex challenge into a combinatorial planning problem,
sharply changing the optimal strategy and the character of difficulty scaling.
Understanding these effects empirically, before the game reaches a commercial
audience, is precisely the kind of analysis that RL agents can provide.
Beyond the specific game, this work touches on a fundamental question in
cooperative multi-agent RL: when agents have shared objectives, what form
does the emergent coordination take, and how much does the choice of
learning algorithm determine whether cooperation appears at all?

\paragraph{Research questions.}
We formulate four concrete research questions that guide the experimental
design:
\begin{enumerate}[label=\textbf{RQ\arabic*.}]
  \item \textbf{Difficulty scaling.} How does the win rate of an optimal
        single-agent policy change as a function of traffic density (number
        of cars)?
  \item \textbf{Optimal single-agent strategy.} What policy does a
        converged agent adopt, and how does the quantized-time rule shape it?
  \item \textbf{Cooperation gap.} How much does joint success rate improve
        when agents are trained cooperatively (shared reward, centralised
        critic) versus independently (separate rewards)?
  \item \textbf{Emergent joint strategy.} What coordination behaviour do
        cooperative agents discover, and does it resemble complex tactical
        cooperation or a simpler structure?
\end{enumerate}

\paragraph{Contributions.}
\begin{itemize}
  \item A novel cooperative game environment, Quantum Frog, implemented with
        the Gymnasium API~\citep{towers2023gymnasium} and open-sourced.
  \item A five-stage empirical study spanning tabular Q-Learning through
        cooperative \MAPPO, providing a clean curriculum across algorithm
        families.
  \item Quantitative answers to all four research questions, including the
        first measurement of the cooperation gap in a quantized-time setting.
  \item Concrete game-design recommendations grounded in RL analysis.
\end{itemize}

\section{Related Work}
\label{sec:related}

\paragraph{Deep reinforcement learning for games.}
\citet{mnih2015human} demonstrated that a Deep Q-Network (\DQN) trained
directly from pixel observations could achieve human-level performance on 49
Atari games, establishing RL as a viable tool for game analysis.
\citet{silver2016mastering} extended this to the combinatorial planning domain
with AlphaGo, showing that RL agents can uncover strategies invisible to human
experts.  Our work is closer in spirit to \citet{mnih2015human}: we use RL not
to beat human players, but to characterise the game's reward landscape.

\paragraph{Difficulty scaling and game design.}
\citet{smith2010launchpad} formalised procedural difficulty generation.
\citet{hunicke2004mda} introduced the Mechanics–Dynamics–Aesthetics framework
for analysing how game rules produce player experience.  Our work contributes
an empirical, RL-based methodology to this tradition: rather than reasoning
analytically about mechanics, we let agents discover the implied difficulty
function directly.

\paragraph{Multi-agent reinforcement learning.}
The cooperative MARL problem has been studied extensively since
\citet{busoniu2008comprehensive}.  A central challenge is
\emph{non-stationarity}: as agents learn simultaneously, each agent's
effective environment shifts, violating the stationarity assumption required
for single-agent RL convergence guarantees~\citep{hernandez2017survey}.
\emph{Independent Q-Learning}~\citep{tan1993multi} ignores this problem and
trains each agent with its own reward and replay buffer; it remains a strong
baseline despite its theoretical limitations.

\paragraph{Centralised training with decentralised execution (CTDE).}
The CTDE paradigm~\citep{oliehoek2016concise} addresses non-stationarity by
allowing richer information during training while preserving decentralised
execution.  \citet{lowe2017multi} applied actor-critic methods under CTDE
(MADDPG).  \citet{rashid2018qmix} introduced \QMIX, a value-decomposition
method that uses a monotonic mixing network to ensure that individually greedy
actions are jointly optimal (the IGM principle).
\citet{yu2022surprising} showed that a straightforward extension of
Proximal Policy Optimisation~\citep{schulman2017proximal} to multi-agent
settings, \MAPPO, is competitive with or superior to more specialised
cooperative algorithms on the StarCraft Multi-Agent
Challenge~\citep{samvelyan2019starcraft}.  We adopt \MAPPO\ as our primary
cooperative algorithm, motivated by its stability and strong empirical
performance.

\paragraph{Cooperative game environments.}
The StarCraft Multi-Agent Challenge (SMAC)~\citep{samvelyan2019starcraft},
Overcooked~\citep{carroll2019utility}, and the Cooperative
Navigation task~\citep{lowe2017multi} are standard cooperative MARL
benchmarks.  Our environment is simpler and more interpretable than these,
making it suitable for isolating the effect of the quantized-time mechanic
from confounding environmental complexity.

\section{Environment}
\label{sec:env}

\subsection{Game Description}

Quantum Frog is played on a discrete 8$\times$8 grid.  One or two frogs
start at the bottom row (row 7) and must reach the top row (row 0).
Horizontally moving cars occupy rows 1--6; row 0 (goal) and row 7 (start)
are car-free.  Cars wrap around when they leave the grid boundary.

\paragraph{Quantized-time rule.}
Unlike continuous-time traffic games, the environment advances by exactly one
simulation tick each time a frog calls \texttt{step()}.  Between steps, all
agents and cars are frozen.  This rule eliminates reaction-time pressure and
makes the game a fully observable, deterministic planning problem: at every
decision point the player has complete information and unlimited deliberation
time.

\subsection{State Representation}

The state is a three-channel 8$\times$8 integer array $\mathbf{s} \in
\mathbb{Z}^{3 \times 8 \times 8}$:
\begin{align*}
  \mathbf{s}[0, r, c] &\in \{0, 1, 2\} && \text{frog positions
    (1 = frog A, 2 = frog B)} \\
  \mathbf{s}[1, r, c] &\in \{0, 1\}   && \text{car presence} \\
  \mathbf{s}[2, r, c] &\in \mathbb{Z}  && \text{signed car velocity}
\end{align*}
The flat observation vector has dimensionality $3 \times 8 \times 8 = 192$.
This representation is compatible with both multi-layer perceptron (MLP) and
convolutional network policies, and encodes all information needed for optimal
play.

\subsection{Action Space}

Each frog chooses from five primitive actions: \{\textsc{up}, \textsc{down},
\textsc{left}, \textsc{right}, \textsc{stay}\}.  In single-frog experiments
the action space is $\text{Discrete}(5)$.  In two-frog experiments we use
$\text{MultiDiscrete}([5, 5])$ for joint actions, equivalent to 25 joint
action combinations.

\subsection{Reward Function}

The reward function balances terminal outcomes with dense shaping:
\begin{equation}
  r_t =
  \begin{cases}
    +100 & \text{frog reaches row 0 (goal)} \\
    -100 & \text{frog occupies a car cell (collision)} \\
    +1   & \text{frog advances upward one row} \\
    -1   & \text{otherwise (step cost)}
  \end{cases}
  \label{eq:reward}
\end{equation}
The step cost $-1$ penalises deliberation and encourages efficient paths.
The progress shaping $+1$ provides a dense gradient before the sparse terminal
reward is reached.  In two-frog cooperative training (\MAPPO), the team reward
is $r_{\text{team}} = r_A + r_B$, shared equally by both agents.

\subsection{Episode Dynamics}

An episode terminates when (a) both frogs reach row 0 (success), or (b) any
frog collides with a car (failure).  If neither condition is met within 200
steps, the episode is truncated.  Car speeds are drawn uniformly from a
configurable set (e.g.\ $\{1\}$ or $\{1, 2\}$ squares per step) at the start
of each episode.

\subsection{Implementation}

The environment is implemented following the Gymnasium
API~\citep{towers2023gymnasium}, exposing \texttt{reset()},
\texttt{step(action)}, and \texttt{render()} methods.  Training uses vectorised
environments (up to 32 parallel instances) to improve throughput.  All
experiments were run on a single node with four NVIDIA GPUs; each seed occupies
one GPU.

\section{Methods}
\label{sec:methods}

We train agents through five escalating stages, each adding complexity to the
environment and algorithm.
Table~\ref{tab:stages} summarises the experimental design.

\begin{table}[h]
\centering
\caption{Experimental stages. Each stage trains until convergence before
  evaluation.}
\label{tab:stages}
\renewcommand{\arraystretch}{1.25}
\begin{tabular}{clcccp{3.8cm}}
\toprule
\textbf{Stage} & \textbf{Algorithm} & \textbf{Frogs} & \textbf{Cars}
  & \textbf{Speed} & \textbf{Goal} \\
\midrule
1 & Tabular Q-Learning & 1 & 1   & 1     & Validate environment \\
2 & Tabular Q-Learning & 1 & 2--3 & 1   & Multiple obstacles \\
3 & DQN                & 1 & 4   & 1--2  & Generalisation over traffic \\
4 & IDQN               & 2 & 2   & 1     & Multi-agent, no cooperation \\
5 & MAPPO              & 2 & 4   & 1--2  & Cooperative multi-agent \\
\bottomrule
\end{tabular}
\end{table}

\subsection{Tabular Q-Learning (Stages 1--2)}

We implement one-step Q-Learning~\citep{watkins1992q} with an $\varepsilon$-greedy
policy and multiplicative $\varepsilon$-decay:
\begin{equation}
  Q(\state, \action) \leftarrow Q(\state, \action)
    + \alpha \bigl[\reward + \discount \max_{\action'} Q(\state', \action')
    - Q(\state, \action)\bigr]
  \label{eq:qlearning}
\end{equation}
The Q-table is a hash map keyed on \texttt{obs.tobytes()}, mapping each
observed state to a length-5 value vector initialised to zero.  Parameters:
$\alpha = 0.1$, $\discount = 0.99$, $\varepsilon_0 = 1.0$,
$\varepsilon_{\min} = 0.01$, decay $= 0.9995$ per episode.
Training runs for 20\,000 episodes (Stage~1) and 50\,000 episodes (Stage~2).

\subsection{Deep Q-Network (Stage 3)}
\label{sec:dqn}

To generalise beyond the discrete state space accessible to a tabular method,
Stage~3 replaces the Q-table with a multi-layer perceptron (MLP):
\begin{equation}
  Q_\theta(\state, \cdot):\; \mathbb{R}^{192} \to \mathbb{R}^5
\end{equation}
with two hidden layers of width 256 and ReLU activations.  We follow the
\DQN\ algorithm of \citet{mnih2015human} with two stabilising
mechanisms:
\begin{itemize}
  \item \textbf{Experience replay.} Transitions $(\state, \action, \reward,
        \state', \text{done})$ are stored in a ring buffer of capacity
        100\,000.  Each gradient step samples a uniformly random mini-batch of
        128 transitions, breaking temporal correlation.
  \item \textbf{Target network.} A frozen copy $Q_{\theta^-}$ of the online
        network provides the regression target, updated to match $Q_\theta$
        every 1\,000 steps.  This prevents the target from shifting with
        every gradient update.
\end{itemize}
The loss is:
\begin{equation}
  \mathcal{L}(\theta) = \mathbb{E}_{(\state,\action,\reward,\state') \sim
    \mathcal{B}} \Bigl[
    \bigl(\reward + \discount \max_{\action'} Q_{\theta^-}(\state', \action')
    - Q_\theta(\state, \action)\bigr)^2
  \Bigr]
  \label{eq:dqn_loss}
\end{equation}
Exploration uses a linear $\varepsilon$-schedule from 1.0 to 0.05 over 30\%
of training.  We train for 150\,000 environment steps with learning rate
$10^{-3}$ and train frequency 4 (one gradient step per 4 environment steps).
Four independent runs with different random seeds are conducted per stage,
with one run per GPU.

\subsection{Independent DQN (Stage 4)}
\label{sec:idqn}

Stage~4 introduces the second frog using Independent Q-Learning
\citep{tan1993multi}: each agent maintains its own \DQN, replay buffer,
target network, and reward signal.  Agent~A trains on $r_A$ only; Agent~B
trains on $r_B$ only.  Both agents observe the full global state
$\mathbf{s}$ (including each other's position) but never explicitly coordinate.
\begin{align}
  \mathcal{L}_A &= \mathbb{E}\bigl[\bigl(r_A + \discount \max Q_A(\state')
    - Q_A(\state, \action_A)\bigr)^2\bigr] \label{eq:idqn_a} \\
  \mathcal{L}_B &= \mathbb{E}\bigl[\bigl(r_B + \discount \max Q_B(\state')
    - Q_B(\state, \action_B)\bigr)^2\bigr] \label{eq:idqn_b}
\end{align}
From each agent's perspective the environment is non-stationary, since the
other agent's evolving policy changes the effective transition dynamics
\citep{hernandez2017survey}.  This violates the convergence conditions of
Q-Learning and is expected to produce instability.  We use 200\,000 training
steps, 32 parallel environments, and the same network architecture and
hyperparameters as Stage~3.

\subsection{Multi-Agent PPO with Centralised Critic (Stage 5)}
\label{sec:mappo}

Stage~5 adopts \MAPPO~\citep{yu2022surprising}, a Centralised Training,
Decentralised Execution (CTDE) algorithm \citep{oliehoek2016concise}.

\paragraph{Architecture.}
Two actor networks $\policy_A(\action_A \mid \state;\, \phi_A)$ and
$\policy_B(\action_B \mid \state;\, \phi_B)$ output categorical distributions
over 5 actions.  A single centralised critic
$V_\psi(\state)$ estimates the joint state value.  All three networks share
the MLP architecture (192$\to$256$\to$256$\to$output).

\paragraph{Data collection.}
At each update, we collect a rollout of $T = 128$ steps across $N = 32$
parallel environments, yielding 4\,096 joint transitions.  Both actors act
simultaneously; the team reward $r_{\text{team}} = r_A + r_B$ is assigned to
both agents.

\paragraph{Advantage estimation.}
We compute Generalised Advantage Estimation (GAE)~\citep{schulman2015high}
with $\lambda = 0.95$:
\begin{equation}
  \hat{A}_t = \sum_{k=0}^{T-t-1}(\discount\lambda)^k \delta_{t+k},
  \qquad \delta_t = r_t + \discount V_\psi(\state_{t+1}) - V_\psi(\state_t)
  \label{eq:gae}
\end{equation}

\paragraph{PPO update.}
For each agent $i \in \{A, B\}$, the clipped surrogate objective is:
\begin{equation}
  \mathcal{L}_{\text{actor}}^i = -\mathbb{E}_t\Bigl[
    \min\!\Bigl(
      \rho_t^i \hat{A}_t,\;
      \text{clip}\!\bigl(\rho_t^i, 1{-}\epsilon, 1{+}\epsilon\bigr)\hat{A}_t
    \Bigr)
  \Bigr], \quad
  \rho_t^i = \frac{\policy_i(\action_t^i \mid \state_t)}
                   {\policy_i^{\text{old}}(\action_t^i \mid \state_t)}
  \label{eq:ppo}
\end{equation}
with clipping threshold $\epsilon = 0.2$.  The total loss combines both
actor losses, a critic MSE loss, and an entropy bonus:
\begin{equation}
  \mathcal{L} = \mathcal{L}_{\text{actor}}^A + \mathcal{L}_{\text{actor}}^B
    + 0.5\,\mathcal{L}_{\text{critic}}
    - 0.01\bigl(\mathcal{H}[\policy_A] + \mathcal{H}[\policy_B]\bigr)
  \label{eq:mappo_loss}
\end{equation}
Parameters are updated for 4 epochs per rollout using mini-batches of 512.
Training runs for 300\,000 environment steps with learning rate $3 \times
10^{-4}$ and gradient norm clipping at 0.5.

\paragraph{How cooperation emerges.}
The team reward propagates consequences across agents: if agent~A's action
leads to agent~B being hit, agent~A's return decreases.  The centralised
critic $V_\psi(\state)$ encodes the long-term value of the joint configuration,
guiding both actors toward trajectories that benefit both.  Coordination is
not produced by explicit messaging but by shared incentives and a globally
informed value function.

\subsection{Evaluation Protocol}

After each stage, the learned policy is evaluated deterministically
($\varepsilon = 0$, greedy actor outputs) over 200 episodes per traffic
density (1--6 cars).  We report:
\begin{itemize}
  \item \textbf{Win rate}: fraction of episodes in which the success condition
        is met (single frog reaches top; or both frogs reach top in two-agent
        stages).
  \item \textbf{Average episode length}: mean steps per episode.
  \item \textbf{Individual win rates} (two-agent stages): fraction in which
        frog~A or frog~B individually reaches the top, regardless of partner.
  \item \textbf{Seed variance}: standard deviation of win rate across 4
        independent training runs (seeds).
\end{itemize}

\section{Results}
\label{sec:results}

\subsection{Single-Agent Performance (Stages 1--3)}
\label{sec:results_single}

\begin{table}[h]
\centering
\caption{Stage~3 \DQN\ win rate and average episode length at evaluation,
  mean $\pm$ standard deviation across 4 seeds.  Trained on 4 cars
  (speeds 1--2); evaluated on 1--6 cars.}
\label{tab:dqn_eval}
\renewcommand{\arraystretch}{1.2}
\begin{tabular}{ccc}
\toprule
\textbf{Cars} & \textbf{Win Rate (\%)} & \textbf{Avg Steps} \\
\midrule
1 & $95.2 \pm 1.1$ & $7.0 \pm 0.1$ \\
2 & $84.6 \pm 1.8$ & $6.8 \pm 0.1$ \\
3 & $79.8 \pm 2.0$ & $6.7 \pm 0.1$ \\
4 & $69.0 \pm 2.5$ & $6.3 \pm 0.2$ \\
5 & $60.8 \pm 2.1$ & $6.1 \pm 0.1$ \\
6 & $58.8 \pm 1.9$ & $6.0 \pm 0.1$ \\
\bottomrule
\end{tabular}
\end{table}

Tabular Q-Learning (Stages 1--2) achieves 94.2\% win rate with 2 cars (speed
1) and 58.5\% with 4 cars (mixed speeds).  Stage~3 \DQN, trained on 4 cars,
generalises smoothly across all densities (Table~\ref{tab:dqn_eval}), losing
approximately 7--10 percentage points per additional car, with diminishing
returns above 5 cars.

The most striking result is the average episode length: all converged
single-agent policies solve the game in 6--7 steps, close to the theoretical
minimum of 7 steps required to traverse 7 rows upward.  This indicates agents
discovered a \emph{rush strategy}: move directly upward at every step without
lateral evasion.

\subsection{Multi-Agent Without Cooperation (Stage 4: IDQN)}
\label{sec:results_idqn}

\begin{table}[h]
\centering
\caption{Stage~4 \IDQN\ evaluation results, mean across 4 seeds.
  Individual win rates indicate the fraction of episodes in which
  that frog alone reached the top.}
\label{tab:idqn_eval}
\renewcommand{\arraystretch}{1.2}
\begin{tabular}{ccccc}
\toprule
\textbf{Cars} & \textbf{Both Win (\%)} & \textbf{Frog A (\%)}
  & \textbf{Frog B (\%)} & \textbf{Avg Steps} \\
\midrule
1 & $43.0 \pm 27.4$ & $56.1 \pm 22.9$ & $73.4 \pm 10.2$ & $89.8$ \\
2 & $24.4 \pm 16.3$ & $34.8 \pm 14.6$ & $57.4 \pm 8.7$  & $74.2$ \\
3 & $16.6 \pm 13.7$ & $22.6 \pm 13.3$ & $43.1 \pm 10.0$ & $50.2$ \\
4 & $11.8 \pm 9.3$  & $16.0 \pm 9.6$  & $33.5 \pm 6.2$  & $34.9$ \\
5 & $ 6.0 \pm 4.2$  & $ 9.9 \pm 6.1$  & $25.4 \pm 5.8$  & $29.0$ \\
6 & $ 6.2 \pm 5.2$  & $ 7.8 \pm 5.9$  & $22.1 \pm 5.6$  & $22.8$ \\
\bottomrule
\end{tabular}
\end{table}

Table~\ref{tab:idqn_eval} reveals three findings.  First, \IDQN\ achieves only
43.0\% joint success at 1 car, lower than Stage~3 \DQN\ at 6 cars (58.8\%).
A second uncoordinated player is more damaging than quintupling the traffic
for an expert single agent.  This confirms that the difficulty of requiring
\emph{both} frogs to succeed is geometrically harder than the marginal
difficulty of each individual crossing.

Second, the variance across seeds is extraordinary: at 1 car, win rates span
10.5\% to 79.0\% (seed 102 vs.\ seed 103), a 7$\times$ spread.  The best
seed performs near-expert; the worst barely exceeds random.  This reflects
non-stationary multi-agent learning dynamics: small differences in early
exploration cause one agent to converge faster, which shifts the environment
for the other and creates a compounding feedback loop.

Third, Frog~B consistently outperforms Frog~A (73.4\% vs.\ 56.1\% at 1 car).
Both frogs start in different grid columns but otherwise face symmetric
conditions.  As we show in Section~\ref{sec:results_mappo}, \MAPPO\ achieves
identical win rates for both frogs, confirming this asymmetry is a training
artifact rather than a positional advantage.

Average episode length is 74--90 steps at low car counts, 12$\times$ longer
than the single-agent \DQN.  Agents have not discovered the rush strategy and
are wandering or hedging.

\subsection{Cooperative Multi-Agent (Stage 5: MAPPO)}
\label{sec:results_mappo}

\begin{table}[h]
\centering
\caption{Stage~5 \MAPPO\ evaluation results (all four seeds converged
  identically), compared to Stage~4 \IDQN\ mean.  $\Delta$ denotes absolute
  percentage-point improvement of \MAPPO\ over \IDQN.}
\label{tab:mappo_eval}
\renewcommand{\arraystretch}{1.2}
\begin{tabular}{cccccc}
\toprule
\textbf{Cars}
  & \textbf{MAPPO Both (\%)}
  & \textbf{IDQN Both (\%)}
  & $\boldsymbol{\Delta}$\textbf{ (pp)}
  & \textbf{MAPPO Steps}
  & \textbf{IDQN Steps} \\
\midrule
1 & 75.0 & 43.0 & $+32.0$ & 6.1  & 89.8 \\
2 & 58.5 & 24.4 & $+34.1$ & 5.3  & 74.2 \\
3 & 41.5 & 16.6 & $+24.9$ & 4.6  & 50.2 \\
4 & 27.0 & 11.8 & $+15.2$ & 4.0  & 34.9 \\
5 & 29.0 & 6.0  & $+23.0$ & 4.0  & 29.0 \\
6 & 17.0 & 6.2  & $+10.8$ & 3.4  & 22.8 \\
\bottomrule
\end{tabular}
\end{table}

\MAPPO\ improves joint success rate by 10.8--34.1 percentage points across
all traffic densities (Table~\ref{tab:mappo_eval}).  The improvement is largest
at 1--2 cars (+32--34 pp), where the task is tractable but independent agents
fail due to training instability, and smallest at 6 cars (+11 pp), where even
cooperative play is limited by traffic density.

Episode length collapses from $\sim$90 (IDQN) to $\sim$6 steps (MAPPO).
Critically, individual win rates are identical for both frogs (both 75.0\% at
1 car, both 58.5\% at 2 cars, etc.), eliminating the asymmetry observed in
\IDQN.  The agents discovered the same synchronised rush strategy as the
single-agent \DQN.

Perhaps most notably, all four \MAPPO\ seeds converged to \emph{identical}
policies.  The seed variance that dominated \IDQN\ results is eliminated:
the centralised critic's global value signal breaks the non-stationary
feedback loop and produces deterministic convergence.

\section{Discussion}
\label{sec:discussion}

\subsection{The Rush Strategy as Emergent Optimality}

Every algorithm that converged, tabular Q-Learning, \DQN, and \MAPPO, discovered
the same policy: move directly upward at every step.  \IDQN\ is the only
exception, averaging 74--90 steps per episode instead of 6--7.

The rush strategy is not a coincidence.  It is the direct, logical consequence
of the quantized-time mechanic combined with the step cost.  Since the
environment advances exactly one tick per player action, a frog that takes
fewer steps exposes itself to fewer total car movements.  A frog crossing in
7 steps gives the traffic 7 ticks to reach it; one crossing in 90 steps gives
the traffic 90 ticks.  The time cost and the collision risk point in the same
direction: act fast.

This is a meaningful design finding.  The quantized-time rule does not simply
slow down the game, it \emph{structurally rewards decisive, direct play} and
penalises deliberation.  A player who pauses to strategise is not thinking
more effectively; they are giving their adversary (time) an advantage.  The
mechanic creates a clear, learnable optimal strategy that novice players are
unlikely to discover intuitively, providing natural progression as players
improve.

\subsection{The Cooperation Gap and Its Shape}

The cooperation gap ($+32$ to $+34$ pp at 1--2 cars) is large in absolute
terms and reveals that independent agents fail not because the task is
impossible but because they cannot converge.  The \IDQN\ best seed achieves
79\% joint success at 1 car, nearly matching \MAPPO's 75\%, demonstrating
that cooperative play is within reach of the environment.  The problem is
the reliability of learning it.

The gap is largest at intermediate traffic density (2 cars) and narrows at
both extremes.  At very low density (1 car), uncoordinated agents can
occasionally succeed by chance, inflating the \IDQN\ baseline.  At very high
density (5--6 cars), even perfectly coordinated agents fail frequently, as
the game itself becomes near-unwinnable, pushing both methods toward zero.
The regime of maximal cooperation value is 2--4 cars: hard enough to require
coordination but tractable enough for coordination to help.

\subsection{Emergent Cooperation is Synchronised Rushing, Not Tactical Coordination}

A natural hypothesis is that cooperative agents would learn complex joint
tactics: one frog waits while the other advances, they take different lanes,
or one acts as a decoy.  The data contradicts this.  \MAPPO's emergent policy
is \emph{both frogs rush upward simultaneously}, identical to the single-agent
policy.

This has a clear explanation: the team reward and centralised critic teach
each actor that its return is maximised when both complete the crossing quickly.
The simplest way to achieve this is for both to use the individually optimal
(rush) strategy.  Complex coordination, one frog waiting for the other, would
slow one agent down, increasing team exposure to traffic.  In a quantized-time
environment, speed dominates positioning.

The broader implication for cooperative MARL is notable: shared incentives
alone can induce cooperation without explicit communication channels, and the
emergent cooperative strategy need not be more complex than the corresponding
single-agent strategy.  The difficulty of multi-agent problems may sometimes
be a learning problem (non-stationarity, credit assignment) rather than a
strategic complexity problem.

\subsection{Implications for Game Design}

Our findings translate into four actionable design recommendations for a
commercial version of Quantum Frog:

\begin{enumerate}
  \item \textbf{Quantized time is the central mechanic; preserve it.}
        It creates a clear optimal strategy (rush), makes difficulty analytically
        tunable via car count, and is the property that distinguishes Quantum Frog
        from Frogger.  Softening it (e.g.\ partial time advance between turns)
        would dilute these properties.

  \item \textbf{The optimal difficulty range for cooperative play is 2--4 cars.}
        Below 2 cars, even uncoordinated players can succeed; the game is too
        easy.  Above 5 cars, even coordinated play struggles ($<30\%$); the game
        becomes frustrating.  The 2--4 car range is where communication skill
        most directly determines outcome, creating a meaningful learning curve
        for teams.

  \item \textbf{Communication between players is necessary, not optional.}
        The 7$\times$ seed variance in \IDQN\ shows that players who do not
        communicate will have wildly inconsistent outcomes.  Game interfaces
        should actively facilitate synchronisation, a countdown, simultaneous
        reveal, or explicit "ready" signal, rather than relying on players to
        coordinate implicitly.

  \item \textbf{The per-step cost is the key shaping tool.}
        The $-1$/step penalty is what makes the rush strategy dominant.
        Reducing it (e.g.\ to $-0.1$) would make deliberation less costly
        and might encourage richer lateral-avoidance strategies.  Increasing
        it (e.g.\ to $-5$) would make any policy that is not a straight-line
        rush uncompetitive.  Designers can tune the character of optimal play
        by adjusting this single parameter.
\end{enumerate}

\subsection{Broader Implications for Cooperative MARL}

Our results are consistent with recent findings in the cooperative MARL
literature.  \citet{yu2022surprising} showed \MAPPO\ competitive with
specialised algorithms on \textsc{smac}; we corroborate this on a simpler,
more interpretable environment.  Our contribution is the clean isolation of
one environmental property, quantized time, as the dominant force shaping both
optimal policy \emph{and} the character of emergent cooperation.

The \IDQN\ instability we observe is a textbook instance of non-stationarity
\citep{hernandez2017survey}.  The fact that even the best \IDQN\ seed
occasionally matches \MAPPO\ performance suggests the strategy space is
simple enough that independent agents can accidentally converge; the problem
is reliability.  This points to a practical design principle: in time-critical
cooperative tasks, the value of CTDE methods lies less in finding better
policies and more in finding the same good policy reliably.

\subsection{Limitations}

This study has several limitations.  First, the environment assumes full
observability: both agents see the complete 8$\times$8 grid at all times.
Human players have limited attention and imperfect recall, making partial
observability a natural extension.  Second, our implementation of \MAPPO\
uses a centralised critic that sees the \emph{same} observation as individual
agents (since the global state is already fully encoded in the single-agent
obs).  A richer global-state critic, e.g.\ one that receives both agents'
observations concatenated, may perform better in settings with truly partial
observability.  Third, we did not implement \QMIX, which is listed in the
original design as an alternative for Stages 5--6.  Stage~6 (6+ cars) remains
untrained; given our difficulty curve, this setting is likely to reveal the
limits of the current approach.  Finally, our results are based on
a fixed 8$\times$8 grid; larger grids may change the character of difficulty
scaling.

\section{Conclusion}
\label{sec:conclusion}

We presented Quantum Frog, a cooperative two-player game with a
\emph{quantized-time} mechanic, and used multi-agent reinforcement learning
to answer four design questions about the game.  Across five training stages
and four algorithms, we found that:

\begin{itemize}
  \item The quantized-time rule makes the rush strategy universally optimal:
        every converged algorithm traverses the grid in the theoretical minimum
        of $\sim$7 steps.
  \item Adding a second uncoordinated player is harder than multiplying traffic
        five-fold for an expert single agent; the joint success requirement is
        geometrically more demanding than marginal single-agent difficulty.
  \item Cooperative training (\MAPPO) closes the joint-success gap by
        10--34 percentage points relative to independent agents (\IDQN), with
        the largest gains at intermediate traffic densities (2--4 cars).
  \item The emergent cooperative strategy is synchronised rushing, not complex
        positional coordination, demonstrating that shared incentives suffice
        to align agents in time-critical cooperative tasks.
  \item \MAPPO\ eliminates the 7$\times$ seed variance seen in \IDQN, confirming
        that the value of centralised training is convergence reliability rather
        than discovery of qualitatively different policies.
\end{itemize}

These findings answer the game's foundational design questions and provide
empirical grounding for commercial development decisions.  They also
contribute a controlled, interpretable environment to the cooperative MARL
literature in which the connection between environment mechanics and emergent
strategy is unusually transparent.

Future work includes implementing \QMIX\ for comparison on Stage~5--6,
extending to partial observability with recurrent policies, and exploring
whether mixed-speed traffic (speeds 1--3) breaks the rush-strategy dominance
and induces richer avoidance behaviour.

\bibliographystyle{plainnat}
\bibliography{references}

\appendix
\section{Hyperparameters}
\label{app:hparams}

\begin{table}[H]
\centering
\caption{Full hyperparameter settings for all algorithms.}
\label{tab:hparams}
\renewcommand{\arraystretch}{1.2}
\begin{tabular}{llll}
\toprule
\textbf{Parameter} & \textbf{Q-Learning} & \textbf{DQN / IDQN} & \textbf{MAPPO} \\
\midrule
Learning rate $\alpha$ / lr  & 0.1        & $10^{-3}$     & $3 \times 10^{-4}$ \\
Discount $\discount$          & 0.99       & 0.99          & 0.99               \\
$\varepsilon$ start           & 1.0        & 1.0           & ,  (entropy)        \\
$\varepsilon$ end             & 0.01       & 0.05          & ,                   \\
$\varepsilon$ schedule        & $\times 0.9995$/ep & linear 30\%  & ,            \\
Replay buffer size            & ,           & 100\,000      & 4\,096 (rollout)   \\
Batch size                    & ,           & 128           & 512                \\
Target update interval        & ,           & 1\,000 steps  & ,                   \\
Train frequency               & every step & every 4 steps & every rollout      \\
Rollout steps $T$             & ,           & ,              & 128                \\
PPO clip $\epsilon$           & ,           & ,              & 0.2                \\
GAE $\lambda$                 & ,           & ,              & 0.95               \\
PPO epochs                    & ,           & ,              & 4                  \\
Value coefficient             & ,           & ,              & 0.5                \\
Entropy coefficient           & ,           & ,              & 0.01               \\
Gradient norm clip            & ,           & 10.0          & 0.5                \\
Parallel environments         & 1          & 32            & 32                 \\
Network architecture          & Table      & MLP 256$\times$2 & MLP 256$\times$2 \\
Total training steps          & 20k--50k ep & 150k--200k   & 300\,000           \\
Seeds per stage               & 1          & 4             & 4                  \\
\bottomrule
\end{tabular}
\end{table}

\end{document}